\documentclass{ieeeaccess}
\usepackage{cite}
\usepackage{amsmath,amssymb,amsfonts}
\usepackage{algorithmic}
\usepackage{graphicx}
\usepackage{textcomp}
\usepackage{subcaption}
\usepackage{multirow}
\usepackage{booktabs}

\def\BibTeX{{\rm B\kern-.05em{\sc i\kern-.025em b}\kern-.08em
    T\kern-.1667em\lower.7ex\hbox{E}\kern-.125emX}}
\begin{document}
\history{Date of publication xxxx 00, 0000, date of current version xxxx 00, 0000.}
\doi{10.1109/ACCESS.2017.DOI}

\title{Low-Field Magnetic Resonance Image Enhancement using Undersampled k-Space}
\author{\uppercase{Daniel Tweneboah Anyimadu}\authorrefmark{1}, 
\uppercase{Mohammed Abdalla\authorrefmark{2}, Mohammed M. Abdelsamea \authorrefmark{1} and Ahmed Karam Eldaly \authorrefmark{1,3}}
\IEEEmembership{Member, IEEE}}
\address[1]{Department of Computer Science, University of Exeter (e-mail: \{da536; m.abdelsamea; a.karam-eldaly\}@exeter.ac.uk)}
\address[2]{Neurology Department, Royal Devon and Exeter Hospital, Exeter, United Kingdom (e-mail: mohamed.abdalla8@nhs.net)}
\address[3]{UCL Hawkes Institute, Department of Computer Science, University College London, London, United Kingdom}

\markboth
{Daniel Tweneboah Anyimadu \headeretal: Low-Field MRI Super-Resolution using Undersampled k-Space}
{Daniel Tweneboah Anyimadu \headeretal: Low-Field MRI Super-Resolution using Undersampled k-Space}

\corresp{Corresponding author: Ahmed K. Eldaly (e-mail: a.karam-eldaly@exeter.ac.uk).}

\begin{abstract}
Low-field magnetic resonance imaging (MRI) offers a cost-effective alternative for medical imaging in resource-limited settings. However, its widespread adoption is hindered by two key challenges: prolonged scan times and reduced image quality. Accelerated acquisition can be achieved using k-space undersampling, while image enhancement traditionally relies on spatial-domain postprocessing. In this work, we propose a novel deep learning framework based on a U-Net variant that operates directly in k-space to super-resolve low-field MR images directly using undersampled data while quantifying the impact of reduced k-space sampling. Unlike conventional approaches that treat image super-resolution as a postprocessing step following image reconstruction from undersampled k-space, our unified model integrates both processes, leveraging k-space information to achieve superior image fidelity. Extensive experiments on synthetic and real low-field brain MRI datasets demonstrate that k-space-driven image super-resolution outperforms conventional spatial-domain counterparts. Furthermore, our results show that undersampled k-space reconstructions achieve comparable quality to full k-space acquisitions, enabling substantial scan-time acceleration without compromising diagnostic utility. \textit{The code will be made publicly available when the manuscript is accepted for publication.}
\end{abstract}


\titlepgskip=-15pt

\maketitle

\section{Introduction}
Magnetic resonance imaging (MRI) enables non-invasive visualisation of anatomical and functional information with high soft-tissue contrast \cite{van2019value}. High-field (HF) MRI systems ($\geq1.5$ T) are the clinical gold standard due to their superior signal-to-noise ratio (SNR), contrast-to-noise ratio (CNR), and spatial resolution \cite{caverly2018rf, vachha2021mri}. However, the high cost and operational complexity of HF-MRI significantly limit accessibility, particularly in low- and middle-income countries and in rural or mobile healthcare settings \cite{marzola2003high}.

\begin{figure}
\centering
\includegraphics[width=0.49\textwidth]{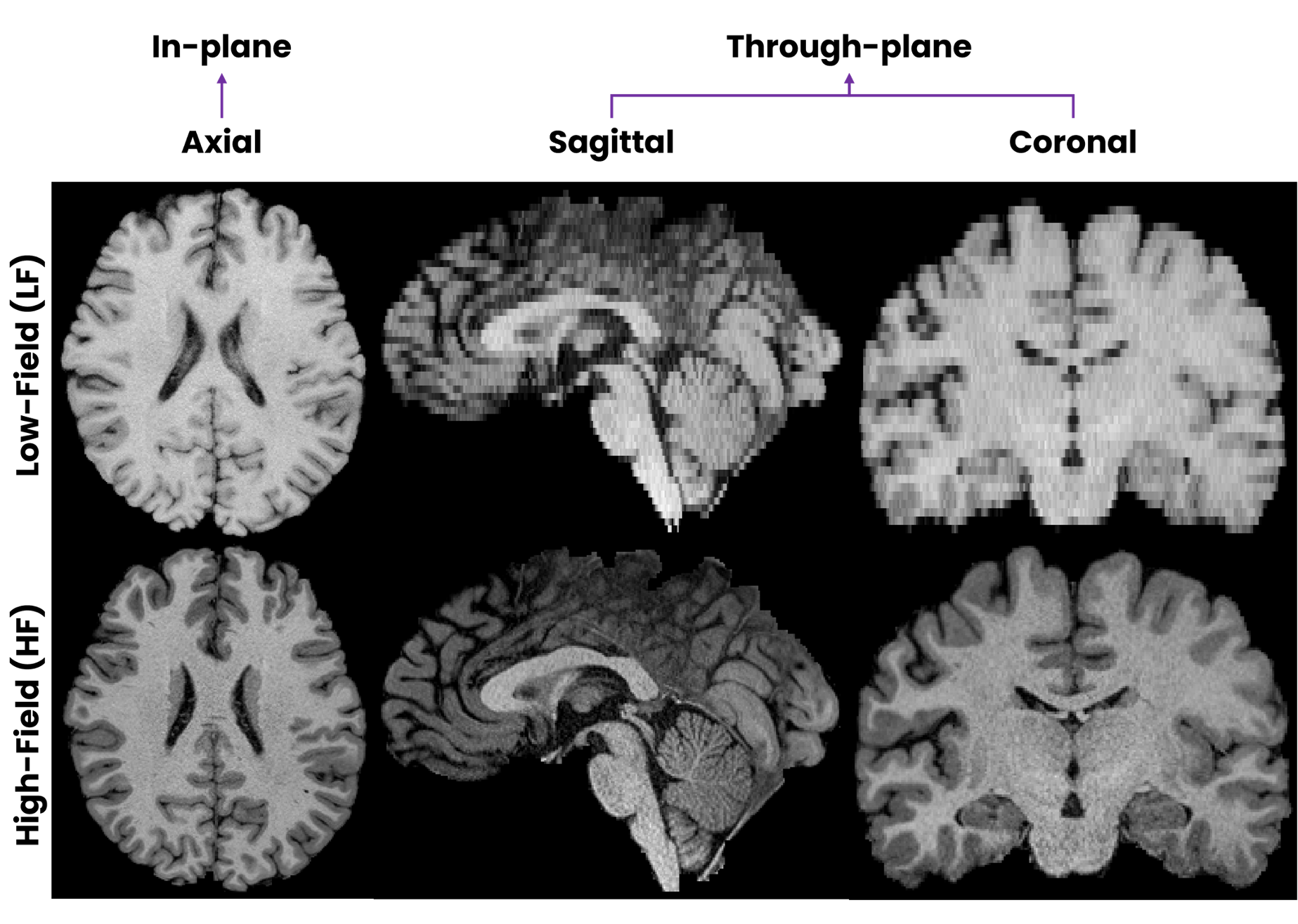}
\caption{Low-field versus high-field MR scans of the same subject. First column shows contrast change on axial plane, and middle and last columns show resolution and contrast change on coronal and sagittal planes. High-field images correspond to the 3T MRI from Human Connectome Project \cite{sotiropoulos2013advances}; and low-field images are simulated from the high-field images using a 0.3T MRI system.}
\label{fig:MRIacquisition}
\end{figure}

Low-field (LF) MRI systems ($<$1 T) have therefore re-emerged as a cost-effective alternative for point-of-care and community-based imaging \cite{arnold2023low, murali2024bringing}. Yet, despite their growing clinical relevance, LF-MRI systems suffer from two persistent and interconnected challenges: (1) prolonged scan times, and (2) reduced image quality, primarily due to the intrinsic limitations imposed by lower SNR \cite{arnold2023low} (as seen in Figure \ref{fig:MRIacquisition}). Moreover, the lack of uncertainty quantification mechanisms compounds the problem, as clinicians are often left without any indication of the reliability of image reconstructions \cite{zou2023review, lambert2022trustworthy}.

To accelerate the acquisition process, k-space undersampling techniques have been employed, thereby reducing scan time \cite{safari2025advancing, singh2023emerging}. Although effective in accelerating acquisition, undersampling leads to an ill-posed inverse problem, and resulting in aliasing, signal loss, and spatial distortion—effects that are particularly exacerbated in the already noise-prone context of LF-MRI \cite{lustig2007sparse}. To address this challenge to enable fast imaging while reconstructing images from undersampled k-space data, various algorithmic solutions have been proposed. Traditional methods include parallel imaging \cite{pruessmann2001advances} and compressed sensing \cite{hamilton2017recent, Eldaly2026ICASSP}, which leverage redundancy in coil sensitivity maps and image sparsity in transform domains, respectively. Recently, deep learning-based approaches have shown tremendous promise by learning complex priors from large datasets, enabling recovery of high-fidelity images even under severe undersampling \cite{safari2025advancing, hong2024complex, Eldaly2024Bayesian, zhu2024advancing, shimron2024accelerating}. However, while these models can effectively reduce artifacts, they often fall short to recover fine anatomical details, a limitation that becomes critical in clinical tasks requiring high spatial precision.

On the other hand, image super-resolution (SR) and image quality transfer (IQT) have been proposed as learning-based strategies to improve the resolution and contrast of LF-MRI scans by learning mappings from high-quality reference scans \cite{alexander2014image, lin2023low, lin2019deep, lin2021generalised, kim20233d, eldaly2024alternative, iglesias2022accurate, iglesias2023synthsr, gopinath2025low, islam2023improving, khateri2025mri, Daniel2026ISBI, Tien2026ISBI}. Both SR and IQT techniques, such as random forest regression \cite{alexander2014image} and self-supervised mixture-of-experts models \cite{lin2021generalised}, enhance spatial resolution and anatomical details in LF-MRI scans, typically by incorporating high-resolution datasets and learning mappings that transfer fine structural information from HF to LF images. For example, recent advancements in synthesising 1 mm isotropic MPRAGE-like scans from LF-MRI T1- and T2-weighted sequences have shown strong correlations with HF measurements, improving the segmentation of brain structures and diagnostic accuracy \cite{iglesias2022accurate}. Tools like ``SynthSR'' convert heterogeneous clinical scans into high-resolution T1-weighted images, improving neuroimaging for diseases like brain tumors and Alzheimer's \cite{iglesias2023synthsr}. Vital as they are, both SR and IQT typically operate in the spatial domain, transforming reconstructed images to enhance their quality by predicting high-frequency details that were lost during acquisition.

On the other hand, the integration of uncertainty quantification into medical image reconstruction, and SR and IQT has garnered increasing attention, driven by the need for transparent, risk-aware AI systems in clinical practice \cite{zou2023review, lambert2022trustworthy, lindenmeyer2025towards, budd2021survey, seoni2023application}. Uncertainty quantification provides a mechanism to estimate the confidence or reliability of each prediction, which is essential for informed clinical decision-making \cite{lakshminarayanan2017simple, shen2022deep}.

To date, existing methods treat LF-MRI reconstruction from undersampled k-space and image SR or IQT as distinct tasks, typically arranged in a sequential pipeline. In such approaches, the raw undersampled k-space data is first reconstructed into an image using a dedicated model. Only after this initial step is the reconstructed image enhanced using SR or IQT techniques. This fragmented approach introduces several critical limitations, particularly in low-field settings where data are inherently noisy and the SNR is low. A key limitation is the loss of frequency-domain information after the initial image is reconstructed. Since the super-resolution/image quality transfer network only operates on spatial-domain data, it no longer has access to the raw spectral features or phase correlations present in k-space—features that are vital for recovering fine anatomical detail. Moreover, the sequential nature of this pipeline leads to error accumulation: artifacts or distortions introduced during reconstruction are passed forward to the enhancement stage, where they can be amplified or misinterpreted as genuine anatomical features. Additionally, applying uncertainty quantification only to the final image fails to capture how uncertainty propagates through each transformation, resulting in uninformative or miscalibrated confidence estimates.

To address these issues, both the reconstruction and image super-resolution/quality transfer stages can be augmented in one model with uncertainty quantification techniques. Thus, in this work, we develop an image super-resolution/image quality transfer technique that operate directly on under-sampled k-space data, rather than relying solely on spatial-domain representations, and thus provide high-quality images akin to those from HF-MRI systems using under-sampled LF-MRI k-space, with quantified uncertainty. Thus, the main contributions of this work are as follows.

\begin{enumerate}
    \item We propose a k-space–based deep learning framework that jointly performs low-field MRI reconstruction, super-resolution/quality transfer, and voxel-wise uncertainty quantification from undersampled k-space data within a single end-to-end model.
    
    \item Unlike existing approaches that treat reconstruction from undersampled k-space and image super-resolution/quality transfer as separate, sequential tasks, the proposed method integrates both processes by operating directly in the frequency domain. This unified formulation enables aggressive k-space undersampling while achieving reconstruction quality comparable to fully sampled high-field MRI acquisitions, consistently outperforming spatial-domain super-resolution/quality transfer methods.
    
    \item The framework incorporates uncertainty quantification to generate voxel-wise confidence maps that capture both undersampling-related degradation and model uncertainty, providing interpretable information on reconstruction reliability.
    
    \item The effectiveness of the proposed approach is demonstrated through extensive experiments on synthetic low-field MRI datasets, including both in-distribution and domain-shifted-out-of-distribution scenarios with available ground truth, as well as on real low-field brain MRI data from healthy and pathological subjects. The evaluation spans multiple undersampling patterns (Cartesian, random, and pseudo-radial) and a range of undersampling ratios, assessing the preservation of clinically relevant image features under aggressive data reduction.
    
    \item We demonstrate that k-space–driven low-field MR image super-resolution/quality transfer enables substantial scan-time acceleration through undersampled k-space acquisition, while maintaining reconstruction quality comparable to fully sampled high-field MRI acquisitions.
\end{enumerate}

\begin{figure*}[t]
\centering
\includegraphics[width=0.9\textwidth]{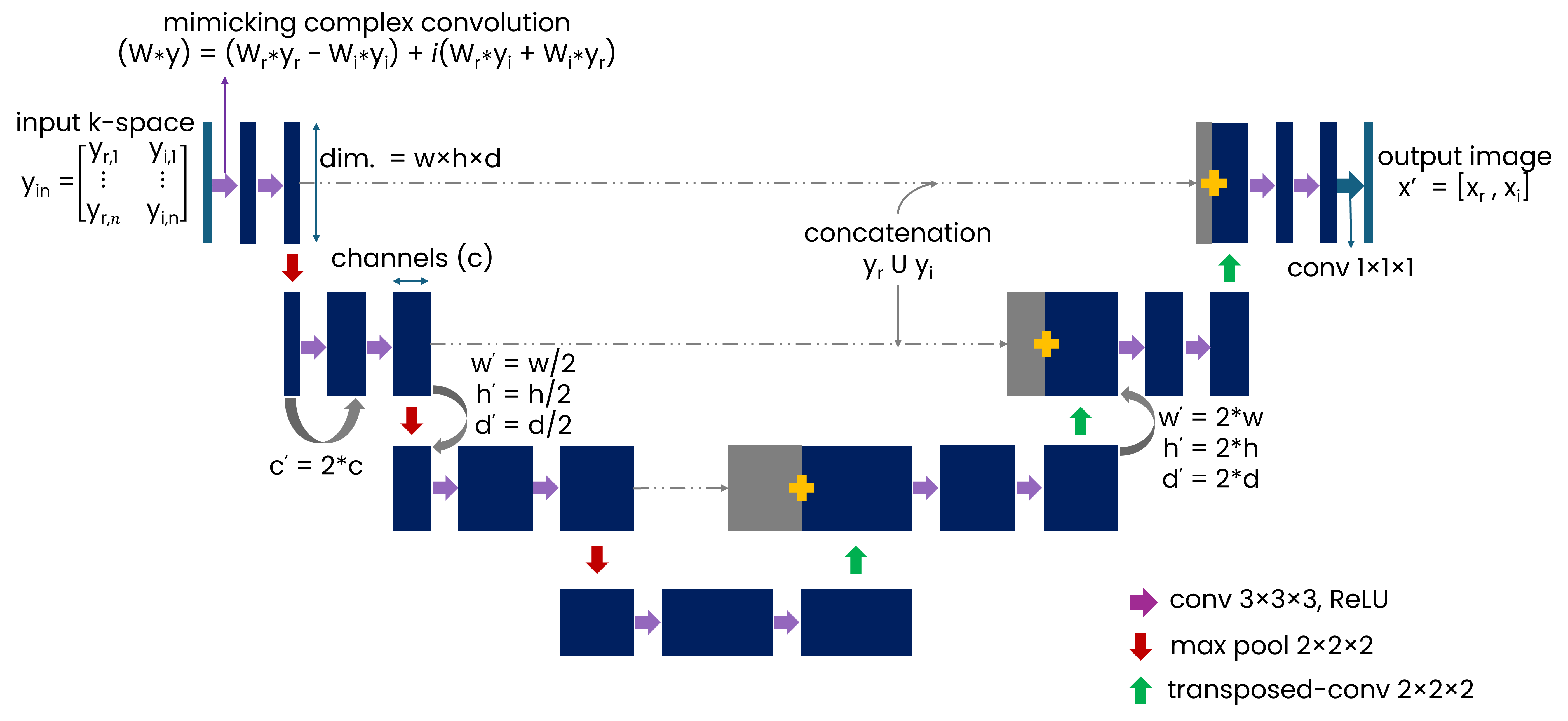}
\caption{kSURF: K-space dual channel U-Net for LF-MRI joint reconstruction, super-resolution/quality transfer and uncertainty quantification from undersampled k-space.}
\label{fig:U-Net}
\end{figure*}

The remaining sections of the paper are organised as follows. Section \ref{sec:Methods} presents the design and implementation details of the proposed approach. Experimental results are then presented in Section \ref{sec:Results}. Finally, conclusions and plans for future work are outlined in Section \ref{sec:Conclusion}.

\section{Methods}
\label{sec:Methods}
This section details the proposed framework and design of our uncertainty-aware deep learning model for LF-MRI reconstruction and super-resolution/quality transfer from under-sampled k-space data. At the core of this framework is a modified 3D U-Net variant architecture \cite{ronneberger2015u}, as illustrated in Figure \ref{fig:U-Net}, designed to process either spatial-domain or frequency-domain (k-space) data representations. To achieve this, the configured architecture employs convolutional operations, activation functions, and interleaved pooling layers to extract both local and global features. These features are progressively refined through a decoder, which reconstructs high-resolution outputs using transposed convolutions and skip connections as in \cite{ronneberger2015u}. The primary distinction between processing spatial-only and k-space data lies in the model’s input and output configurations, as well as how each convolutional filter processes magnitude-only data for the spatial domain and both real and imaginary components for the k-space domain. To highlight these distinctions more clearly, the spatial-domain architecture operates on magnitude-only MRI data, where the convolutional layers process only the magnitude information, which represents the voxel intensities of the data. Since the input consists of a single channel (i.e., the magnitude of the MRI signal), the output is also a single channel, representing the reconstructed high-fidelity spatial data. In contrast, the k-space architecture, referred to as the \textit{k-space dual channel U-Net}, is designed to process raw k-space data directly. This model treats the real and imaginary components of each voxel as separate real-valued channels, stored as floating-point tensors for convolutional processing. Convolutions are then applied to approximate complex-valued convolutions. Mathematically, this is expressed as
\begin{equation}
\quad (\mathbf{W} \ast \mathbf{y}) = (\mathbf{W_r} \ast \mathbf{y_r} - \mathbf{W_i} \ast \mathbf{y_i}) + i(\mathbf{W_r} \ast \mathbf{y_i} + \mathbf{W_i} \ast \mathbf{y_r}),
\end{equation}
where \(\mathbf{W_r}, \mathbf{W_i}\) are the real and imaginary parts of the convolutional weights, respectively, and \(\mathbf{y_r}, \mathbf{y_i}\) are the corresponding input observations. This formulation ensures that both magnitude and phase information are preserved throughout the reconstruction process.

To achieve HF-MRI like reconstruction and enhancement from undersampled low-field data, we train both our spatial-domain and k-space dual-channel U-Net models using paired input-output examples. These pairs are generated by transforming HF-MRI data into synthetic low-field counterparts using a stochastic simulator \cite{lin2019deep, lin2023low}. The corresponding k-space data are then under-sampled at varying acceleration rates, using different undersampling patterns (e.g., pseudo-radial, Cartesian, and random), to simulate MRI acquisitions under accelerated conditions. This paired dataset, comprising under-sampled low-field data and fully sampled high-field data, includes both spatial-domain and k-space representations. It enables the networks to learn accurate mappings for reconstruction in both domains. Specifically: the spatial-domain model learns to map low-field input to high-field output in the image space, whereas the k-space dual-channel model learns to map low-field k-space data to the high-field k-space domain. In addition, the models enhance image quality by recovering spatial-domain and frequency-domain components that are lost during the undersampling process. Once trained, the models are capable of reconstructing high-field-like MR images from undersampled low-field MR spatial domain images and k-space inputs. The loss function for training incorporates two key components: reconstruction accuracy, measured using mean squared error (MSE) and mean absolute error (MAE) between the predicted high-field output \( x'_i \) and the actual high-field reference \( x_i \)
\begin{equation}
\mathcal{L}_{MSE} = \frac{1}{N} \sum_{i=1}^{N} \left\| x'_i - x_i \right\|_2^2, \quad \mathcal{L}_{MAE} = \frac{1}{N} \sum_{i=1}^{N} \left| x'_i - x_i \right|.
\end{equation}

These losses penalise discrepancies between predicted and true values, improving reconstruction accuracy and robustness. Regularisation is applied through weight decay (L2 regularisation) in the optimiser, with the regularisation term defined as
\begin{equation}
\mathcal{L}_{L2} = \lambda \sum_{w} w^2,
\end{equation}
where \( w \) represents the model weights and \( \lambda \) is the weight decay coefficient. The total loss function combines MSE, MAE, and L2 regularisation:
\begin{equation}
\mathcal{L}_{total} = \mathcal{L}_{MSE} + \mathcal{L}_{MAE} + \mathcal{L}_{L2},
\end{equation}
ensuring that the model optimises for both reconstruction accuracy and generalisation.

In addition to the reconstruction loss, we implement an ensemble approach using cross-validation, as in \cite{lakshminarayanan2017simple, dutschmann2023large}, where multiple models are trained on different subsets of the data. Their predictions are then aggregated to estimate uncertainty through variance. This uncertainty estimation provides a robust and reliable assessment of the model's confidence in the reconstructed MR images. The uncertainty in the output \( x'_i \) is quantified by the variance of the predictions across different models, given as: $\mathcal{U}(x'_i) = \text{Var}(x'_i)$.

\subsection{Deep learning architecture configuration}
\label{ssec:arch}
Figure \ref{fig:U-Net} illustrates our modified three-layer U-Net \cite{ronneberger2015u}, a k-space dual-channel architecture designed for reconstructing complex-valued MRI k-space. The input of the network is represented as two real-valued channels and is stored as floating point tensors, corresponding to the real $\mathcal{R}(\cdot)$ and imaginary $\mathcal{I}(\cdot)$ components of the k-space. The network learns a two-channel nonlinear mapping, enabling it to approximate complex arithmetic while preserving both magnitude and phase information, ultimately generating a high-field-like k-space output in the same format. This mapping can be expressed mathematically as
\begin{equation}
\hat{\mathbf{y}}_{\mathrm{HF}} = f_{\theta}\!\bigl(\mathcal{R}(\mathbf{y}_{\mathrm{LF}}^{\mathrm{us}}),\, \mathcal{I}(\mathbf{y}_{\mathrm{LF}}^{\mathrm{us}})\bigr),
\end{equation}
where $f_{\theta}$ represents our k-space dual-channel U-Net architecture. To maintain a consistent baseline for comparing models trained in both the spatial and k-space domains, we utilise standard convolutional layers. Notably, the encoder of the k-space dual-channel network consists of three convolutional blocks (convBlock), each containing two consecutive $3 \times 3 \times 3$ convolutional layers followed by ReLU activations, and interleaved with $2 \times 2 \times 2$ max-pooling layers for downsampling. Specifically, the first block outputs 64 channels, the second 128 channels, and the third 256 channels. At the bottleneck, two $3 \times 3 \times 3$ convolutional layers with 512 output channels capture the most abstract representations of the input data. The decoder mirrors the encoder with three upsampling stages using $2 \times 2 \times 2$ transposed convolutions. Each stage upsamples the feature maps and concatenates them with the corresponding encoder block via skip connections. The first decoder block upsamples from 512 to 256 channels and applies a convBlock with two $3 \times 3 \times 3$ convolutions; the second upsamples from 256 to 128 channels with a similar convBlock; and the third upsamples from 128 to 64 channels. Finally, a $1 \times 1 \times 1$ convolutional layer outputs two channels representing the reconstructed real and imaginary k-space components. All convolutional layers use ReLU activations, except for the final output layer. 

\subsection{Experiments}
\subsubsection{Datasets for model training}
Our primary HF-MRI dataset is sourced from the WU-Minn Human Connectome Project (HCP) \cite{Sotiropoulos2013Oct}, consisting of three-dimensional, high-resolution T1-weighted (T1w) brain volumes acquired on a 3T Siemens Connectome Skyra scanner, with 0.7-mm isotropic voxel resolution. The T1w images were acquired using the following imaging parameters: repetition time (TR) of 2400 ms, echo time (TE) of 2.14 ms, and inversion time (TI) of 1000 ms. For model development, 15 volumes were randomly selected from the HCP dataset to ensure anatomical diversity for training, while 50 volumes were used for testing to comprehensively evaluate the model’s performance. To simulate LF-MRI data, we employed a stochastic low-field image simulator that converts high-field images into realistic low-field counterparts \cite{lin2019deep, lin2023low}. This process models tissue-specific contrast variations by adjusting the SNR in gray matter (GM) and white matter (WM), ensuring alignment with the characteristics of lower magnetic field strength imaging \cite{lin2023low}. The in-distribution (InD) dataset used for training is synthesised with parameters constrained by a Mahalanobis distance $< 1$, ensuring that the SNR in white matter (WM) is higher than in gray matter (GM) to maintain tissue contrast compatible with T1-weighted images as in \cite{lin2019deep, lin2023low}. The simulation mimics the degraded contrast and reduced SNR typically observed in clinical images generated by LF-MRI systems as seen in Figure \ref{fig:MRIacquisition}.

\subsubsection{Undersampling strategies}
\begin{figure}
    \centering
    \includegraphics[width=0.4\textwidth]{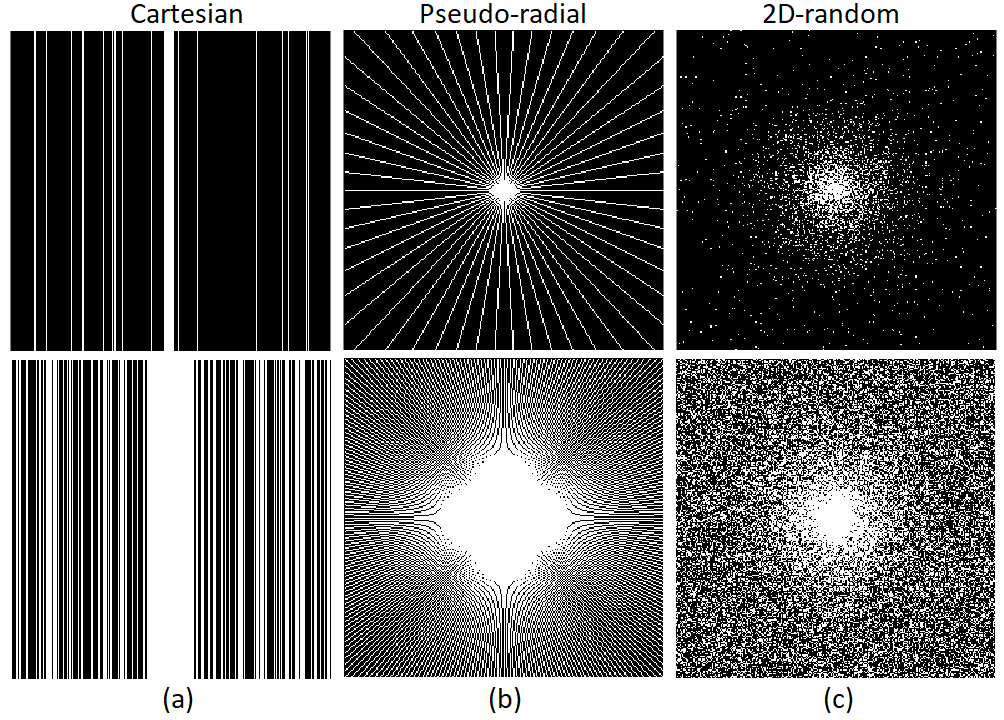}
    \caption{Images of sampling masks. (a) Cartesian sampling,(b) pseudo-radial sampling and (c) 2D variable density random sampling. Here, we show the 10\% (top) and 50\% (bottom) masks for Cartesian and Pseudo-radial sampling, and 5\% (top) and 40\% (bottom) masks of 2D-random sampling.}
    \label{fig:Sampling}
\end{figure}

To emulate accelerated MRI acquisition on the simulated low-field k-space volumes from the HCP data, we utilised three widely used patterns in the literature: 2D-random, Cartesian with random phase encodes (1D random), and pseudo-radial undersampling patterns \cite{yao2018efficient, xie2020iterative}. While Cartesian and pseudo-radial sampling patterns are commonly used in actual MRI acquisition protocols, the 2D-random sampling is not used in practice but provides a useful theoretical baseline in evaluating new magnetic resonance image reconstruction methodologies. In our framework, the undersampling ratios for Cartesian and pseudo-radial sampling were set to {10\%, 20\%, 30\%, 40\%, 50\%}, while for 2D-random sampling, the ratios were set to {5\%, 10\%, 20\%, 30\%, 40\%}, including fully sampled 100\%. Figure \ref{fig:Sampling} shows example masks from the three subsampling strategies.

\subsubsection{Data preprocessing}
Prior to model training, the generated paired low-field and high-field MRI volumes were spatially cropped to a standardised region with dimensions of 256 (axial) $\times$ 288 (coronal) $\times$ 256 (sagittal) voxels and the low-field images were interpolated in depth to match high-field slice counts, ensuring volumetric alignment. Each volume was transformed to complex-valued k-space using centered 3D fast Fourier transform (FFT) \cite{brigham1988fast, hansen2015image}, followed by application of undersampling masks to both the high-field and the low-field volumes to generate k-space data. Under-sampled k-space was converted back to image space via inverse FFT, from which overlapping 3D patches of size $32 \times 32 \times 32$ voxels were extracted with a stride of 16 voxels. Corresponding high-field patches were extracted from the fully sampled high-field images at the same spatial locations to form paired training examples. These image patches were then converted back to k-space by FFT, and their real and imaginary components separated and stacked as two-channel inputs for the network. The patch-based dataset was aggregated into arrays, reshaped for efficient batch processing, and used as the training dataset.

\subsubsection{Implementation and Training}
Our framework was implemented in Python \cite{van2007python} using PyTorch \cite{paszke2019pytorch} and trained on an NVIDIA Tesla A100 graphical processing unit (GPU) with 80GB video random access memory (VRAM). The model was optimised using the Adam optimiser with a learning rate of \(1 \times 10^{-4}\) and weight decay of \(1 \times 10^{-6}\). Training was conducted with a batch size of 16 over 100 epochs, ensuring convergence of the validation loss. The dataset was subjected to three-fold cross-validation, where it was split into three subsets to monitor performance, prevent overfitting and ensure generalisability. The loss function combined MSE \cite{Prasad01031990}, and MAE \cite{Willmott2005AdvantagesOT}, to balance voxel-wise accuracy and edge preservation. Training resumed from existing checkpoints when available, ensuring continuity. Validation was conducted to assess reconstruction fidelity, with the best model checkpoint saved based on the lowest validation MSE. Training progress was monitored at each epoch to ensure convergence and consistent improvement.

\subsubsection{Testing}
We evaluated our method on four datasets: two synthetic LF-MRI datasets derived from the HCP and two publicly available real low-field datasets described in \cite{wogu2025labeled}. The HCP-derived test sets include an in-distribution (InD) dataset and an out-of-distribution (OOD) dataset, each comprising 50 T1-weighted volumes from 50 distinct subjects. The InD dataset was synthesised using imaging parameters sampled from the same distribution as the training data as described earlier. In contrast, the OOD dataset was generated by sampling SNR parameters from a distribution representative of ultra–low-field T1-weighted MRI, thereby introducing substantial deviations from the Gaussian parameter distribution used during training. For real low-field evaluation, we used data acquired on Hitachi AIRIS II 0.3T scanners equipped with a 12-channel head coil \cite{wogu2025labeled}. We selected subjects from both healthy and clinically affected cohorts, focusing on T1-weighted images to further assess model robustness and reconstruction quality. The representative healthy subject was a 15-year-old female and the unhealthy subject a 55-year-old male with Parkinson’s disease, both from the South-South region of Nigeria with middle socioeconomic status. All four datasets underwent the same preprocessing pipeline applied during training, including cropping, Fourier transformation, k-space undersampling application, and patch extraction. Quantitative evaluation was performed on the two synthetic datasets for which high-field ground truth volumes are available. We report the structural similarity index measure (SSIM) \cite{Brunet2012OnTM} and peak signal-to-noise ratio (PSNR) \cite{Willmott2005AdvantagesOT}, computed between the reconstructed images and the fully sampled high-field references, for the low-field MR/zero-filling images, reconstructions using both spatial domain (sIQT) and k-space super-resolution (kSURF). SSIM quantifies structural preservation, whereas PSNR measures overall voxel-wise fidelity.

\section{Results}
\label{sec:Results}
\subsection{Quantitative Results using Synthetic LF-MRI Datasets}
Tables \ref{tab:InD1pcr} and \ref{tab:OODpcr} report quantitative reconstruction performance in terms of PSNR and SSIM for InD and OOD synthetic LF-MRI datasets, respectively, comparing LF/zero-filling, sIQT, and kSURF under pseudo-radial, Cartesian, and random undersampling patterns across increasing undersampling rates. As expected, baseline LF/zero-filling reconstructions consistently exhibit lower SSIM and PSNR values, reflecting substantial loss of anatomical detail and perceptual fidelity, with degradation becoming increasingly pronounced as the undersampling rate increases. This behaviour confirms the inherent limitations of direct zero-filling reconstruction under reduced k-space sampling. In contrast, both sIQT and kSURF substantially enhance reconstruction quality across all sampling patterns and undersampling rates, yielding marked improvements in both SSIM and PSNR. While reconstruction performance decreases for all methods as undersampling becomes more severe, kSURF and sIQT demonstrate significantly greater robustness compared to LF/zero-filling, maintaining higher SSIM and PSNR values in both InD and OOD datasets. This trend indicates improved generalisation under domain shift, with kSURF consistently exhibiting the strongest resilience to changes in data distribution. Across all sampling schemes and undersampling levels, kSURF achieves the highest quantitative performance, highlighting the effectiveness of k-space–driven super-resolution in preserving essential frequency-domain information and enabling superior structural preservation and voxel-level fidelity, even under aggressive undersampling. Finally, reconstruction quality is influenced by the choice of undersampling pattern, with pseudo-radial sampling consistently yielding the highest SSIM and PSNR values, followed by random and Cartesian sampling, which can be attributed to the more uniform k-space coverage and increased incoherence provided by pseudo-radial masks.

\begin{table*} 
\centering
\begin{subtable}{\textwidth}
\caption{Average SSIM and PSNR, for kSURF and sIQT domains at increasing sampling rates using InD dataset. Bold indicates best results.}
\centering
\label{tab:InD1pcr}
\resizebox{0.95\textwidth}{!}{
\begin{tabular}{c|c|c|c|c|c|c|c}
\hline\hline
\multicolumn{2}{l}{\textbf{Pseudo-radial Sampling}} \\
\toprule
\multicolumn{2}{c|}{\textbf{Domain / Metrics}} & \textbf{100\%} & \textbf{50\%} & \textbf{40\%} & \textbf{30\%} & \textbf{20\%} & \textbf{10\%} \\
\midrule\hline
\multirow{3}{*}{\textbf{LF/Zero-filling}} 
& SSIM $\uparrow$ & 0.90 $\pm$ 0.01 & 0.63 $\pm$ 0.07 & 0.57 $\pm$ 0.07 & 0.52 $\pm$ 0.07 & 0.44 $\pm$ 0.05 & 0.35 $\pm$ 0.03 \\
& PSNR $\uparrow$ & 29.95 $\pm$ 2.87 & 24.50 $\pm$ 2.34 & 24.49 $\pm$ 2.24 & 24.36 $\pm$ 2.24 & 24.06 $\pm$ 1.57 & 22.94 $\pm$ 0.84 \\
\midrule\hline
\multirow{3}{*}{\textbf{sIQT}} 
& SSIM $\uparrow$ & 0.96 $\pm$ 0.01 & 0.96 $\pm$ 0.01 & 0.95 $\pm$ 0.01 & 0.95 $\pm$ 0.01 & 0.94 $\pm$ 0.01 & 0.93 $\pm$ 0.02 \\
& PSNR $\uparrow$ & 35.55 $\pm$ 2.54 & 35.15 $\pm$ 2.57 & 34.85 $\pm$ 2.72 & 34.42 $\pm$ 2.65 & 33.56 $\pm$ 2.67 & 31.06 $\pm$ 2.71 \\
\midrule\hline
\multirow{3}{*}{\textbf{kSURF}} 
& SSIM $\uparrow$ & \textbf{0.97 $\pm$ 0.01} & \textbf{0.97 $\pm$ 0.01} & \textbf{0.97 $\pm$ 0.01} & \textbf{0.97 $\pm$ 0.01} & \textbf{0.96 $\pm$ 0.01} & \textbf{0.94 $\pm$ 0.01} \\
& PSNR $\uparrow$ & \textbf{35.60 $\pm$ 2.45} & \textbf{35.18 $\pm$ 2.50} & \textbf{35.14 $\pm$ 2.57} & \textbf{34.62 $\pm$ 2.72} & \textbf{33.70 $\pm$ 2.70} & \textbf{31.21 $\pm$ 2.55} \\

\hline\hline
\multicolumn{2}{l}{\textbf{Cartesian Sampling}} \\
\midrule\hline
\multicolumn{2}{c|}{\textbf{Domain / Metrics}} & \textbf{100\%} & \textbf{50\%} & \textbf{40\%} & \textbf{30\%} & \textbf{20\%} & \textbf{10\%} \\
\midrule
\multirow{3}{*}{\textbf{LF/Zero-filling}} 
& SSIM $\uparrow$ & 0.90 $\pm$ 0.01 & 0.72 $\pm$ 0.03 & 0.70 $\pm$ 0.03 & 0.68 $\pm$ 0.03 & 0.65 $\pm$ 0.03 & 0.60 $\pm$ 0.03 \\
& PSNR $\uparrow$ & 29.95 $\pm$ 2.87 & 23.94 $\pm$ 1.77 & 23.88 $\pm$ 1.59 & 23.41 $\pm$ 1.37 & 22.80 $\pm$ 0.90 & 20.42 $\pm$ 0.64 \\
\midrule\hline
\multirow{3}{*}{\textbf{sIQT}} 
& SSIM $\uparrow$ & 0.96 $\pm$ 0.01 & 0.95 $\pm$ 0.01 & 0.94 $\pm$ 0.01 & 0.93 $\pm$ 0.02 & 0.92 $\pm$ 0.02 & 0.90 $\pm$ 0.02 \\
& PSNR $\uparrow$ & 35.55 $\pm$ 2.54 & 33.86 $\pm$ 2.46 & 33.14 $\pm$ 2.66 & 31.81 $\pm$ 2.62 & 30.70 $\pm$ 2.77 & 28.20 $\pm$ 1.98 \\
\midrule\hline
\multirow{3}{*}{\textbf{kSURF}} 
& SSIM $\uparrow$ & \textbf{0.97 $\pm$ 0.01} & \textbf{0.96 $\pm$ 0.01} & \textbf{0.96 $\pm$ 0.01} & \textbf{0.95 $\pm$ 0.01} &  \textbf{0.94 $\pm$ 0.01} & \textbf{0.92 $\pm$ 0.01} \\
& PSNR $\uparrow$ & \textbf{35.60 $\pm$ 2.45} & \textbf{33.95 $\pm$ 2.47} & \textbf{33.30 $\pm$ 2.54} & \textbf{32.10 $\pm$ 2.57} & \textbf{30.85 $\pm$ 2.53} & \textbf{28.35 $\pm$ 2.27} \\

\hline\hline
\multicolumn{2}{l}{\textbf{2D Random Sampling}} \\
\midrule\hline
\multicolumn{2}{c|}{\textbf{Domain / Metrics}} & \textbf{100\%} & \textbf{40\%} & \textbf{30\%} & \textbf{20\%} & \textbf{10\%} & \textbf{5\%} \\
\midrule\hline
\multirow{3}{*}{\textbf{LF/Zero-filling}} 
& SSIM $\uparrow$ & 0.90 $\pm$ 0.01 & 0.44 $\pm$ 0.06 & 0.40 $\pm$ 0.05 & 0.36 $\pm$ 0.04 & 0.33 $\pm$ 0.03 & 0.29 $\pm$ 0.03 \\
& PSNR $\uparrow$ & 29.95 $\pm$ 2.87 & 24.21 $\pm$ 1.88 & 23.96 $\pm$ 1.65 & 23.61 $\pm$ 1.48 & 23.03 $\pm$ 1.19 & 22.10 $\pm$ 0.85 \\
\midrule\hline
\multirow{3}{*}{\textbf{sIQT}} 
& SSIM $\uparrow$ & 0.96 $\pm$ 0.01 & 0.95 $\pm$ 0.01 & 0.95 $\pm$ 0.01 & 0.95 $\pm$ 0.01 & 0.94 $\pm$ 0.02 & 0.93 $\pm$ 0.02 \\
& PSNR $\uparrow$ & 35.55 $\pm$ 2.54 & 34.77 $\pm$ 2.56 & 34.17 $\pm$ 2.65 & 33.49 $\pm$ 2.59 & 32.56 $\pm$ 2.74 & 31.16 $\pm$ 2.53 \\
\midrule\hline
\multirow{3}{*}{\textbf{kSURF}} 
& SSIM $\uparrow$ & \textbf{0.97 $\pm$ 0.01} & \textbf{0.97 $\pm$ 0.01} & \textbf{0.97 $\pm$ 0.01} & \textbf{0.96 $\pm$ 0.01} & \textbf{0.96 $\pm$ 0.01} & \textbf{0.95 $\pm$ 0.01} \\
& PSNR $\uparrow$ & \textbf{35.60 $\pm$ 2.45} & \textbf{35.00 $\pm$ 2.52} & \textbf{34.20 $\pm$ 2.57} & \textbf{33.63 $\pm$ 2.61} & \textbf{32.72 $\pm$ 2.68} & \textbf{31.64 $\pm$ 2.69} \\
\bottomrule
\end{tabular}
}
\end{subtable}

\vspace{1em}

\begin{subtable}{\textwidth}
\centering
\caption{Average SSIM and PSNR, for kSURF and sIQT domains at increasing sampling rates using OOD dataset. Bold indicates best results.}
\centering
\label{tab:OODpcr}
\resizebox{0.95\textwidth}{!}{
\begin{tabular}{c|c|c|c|c|c|c|c}
\hline\hline
\multicolumn{2}{l}{\textbf{Pseudo-radial Sampling}} \\
\toprule
\multicolumn{2}{c|}{\textbf{Domain / Metrics}} & \textbf{100\%} & \textbf{50\%} & \textbf{40\%} & \textbf{30\%} & \textbf{20\%} & \textbf{10\%} \\
\midrule\hline
\multirow{3}{*}{\textbf{LF/Zero-filling}} 
& SSIM $\uparrow$ & 0.84 $\pm$ 0.02 & 0.46 $\pm$ 0.09 & 0.42 $\pm$ 0.08 & 0.39 $\pm$ 0.07 & 0.35 $\pm$ 0.06 & 0.30 $\pm$ 0.04 \\
& PSNR $\uparrow$ & 26.39 $\pm$ 3.01 & 20.41 $\pm$ 1.60 & 20.39 $\pm$ 1.69 & 20.36 $\pm$ 1.44 & 20.31 $\pm$ 1.19 & 20.09 $\pm$ 0.77 \\
\midrule\hline
\multirow{3}{*}{\textbf{sIQT}} 
& SSIM $\uparrow$ & 0.88 $\pm$ 0.02 & 0.88 $\pm$ 0.02 & 0.88 $\pm$ 0.03 & 0.88 $\pm$ 0.03 & 0.87 $\pm$ 0.03 & 0.86 $\pm$ 0.02 \\
& PSNR $\uparrow$ & 30.21 $\pm$ 3.50 &  26.16 $\pm$ 2.20 & 26.01 $\pm$ 1.99 & 25.35 $\pm$ 2.21 & 25.12 $\pm$ 2.05 & 23.90 $\pm$ 2.10 \\
\midrule\hline
\multirow{3}{*}{\textbf{kSURF}} 
& SSIM $\uparrow$ & \textbf{0.89 $\pm$ 0.03} & \textbf{0.89 $\pm$ 0.03} & \textbf{0.89 $\pm$ 0.03} & \textbf{0.89 $\pm$ 0.03} & \textbf{0.88 $\pm$ 0.03} & \textbf{0.86 $\pm$ 0.03} \\
& PSNR $\uparrow$ & \textbf{30.30 $\pm$ 3.48} & \textbf{26.30 $\pm$ 1.96} & \textbf{26.24 $\pm$ 2.02} & \textbf{25.60 $\pm$ 2.09} & \textbf{25.20 $\pm$ 2.12} & \textbf{24.01 $\pm$ 1.79} \\

\hline\hline
\multicolumn{2}{l}{\textbf{Cartesian Sampling}} \\
\midrule\hline
\multicolumn{2}{c|}{\textbf{Domain / Metrics}} & \textbf{100\%} & \textbf{50\%} & \textbf{40\%} & \textbf{30\%} & \textbf{20\%} & \textbf{10\%} \\
\midrule\hline
\multirow{3}{*}{\textbf{LF/Zero-filling}} 
& SSIM $\uparrow$ & 0.84 $\pm$ 0.02 & 0.64 $\pm$ 0.04 & 0.63 $\pm$ 0.04 & 0.62 $\pm$ 0.03 & 0.60 $\pm$ 0.03 & 0.58 $\pm$ 0.03 \\
& PSNR $\uparrow$ & 26.39 $\pm$ 3.01 & 20.41 $\pm$ 1.60 & 20.16 $\pm$ 1.16 & 20.10 $\pm$ 1.05 & 19.79 $\pm$ 0.83 & 18.63 $\pm$ 0.59 \\
\midrule\hline
\multirow{3}{*}{\textbf{sIQT}} 
& SSIM $\uparrow$ & 0.88 $\pm$ 0.02 & 0.88 $\pm$ 0.03 & 0.87 $\pm$ 0.03 & 0.86 $\pm$ 0.03 & 0.86 $\pm$ 0.02 & 0.85 $\pm$ 0.02 \\
& PSNR $\uparrow$ & 30.21 $\pm$ 3.50 & 25.95 $\pm$ 1.89 &  25.20 $\pm$ 2.00 & 24.90 $\pm$ 1.84 & 24.34 $\pm$ 1.83 & 23.60 $\pm$ 1.36 \\
\midrule\hline
\multirow{3}{*}{\textbf{kSURF}} 
& SSIM $\uparrow$ & \textbf{0.89 $\pm$ 0.03} & \textbf{0.89 $\pm$ 0.03} & \textbf{0.88 $\pm$ 0.03} & \textbf{0.87 $\pm$ 0.03} & \textbf{0.87 $\pm$ 0.03} & \textbf{0.85 $\pm$ 0.02} \\
& PSNR $\uparrow$ & \textbf{30.30 $\pm$ 3.48} & \textbf{26.06 $\pm$ 1.73} & \textbf{25.30 $\pm$ 2.08} & \textbf{24.95 $\pm$ 1.81} & \textbf{24.38 $\pm$ 1.60} & \textbf{24.07 $\pm$ 1.43} \\

\hline\hline
\multicolumn{2}{l}{\textbf{2D Random Sampling}} \\
\midrule\hline
\multicolumn{2}{c|}{\textbf{Domain / Metrics}} & \textbf{100\%} & \textbf{40\%} & \textbf{30\%} & \textbf{20\%} & \textbf{10\%} & \textbf{5\%} \\
\midrule\hline
\multirow{3}{*}{\textbf{LF/Zero-filling}} 
& SSIM $\uparrow$ & 0.84 $\pm$ 0.02 & 0.34 $\pm$ 0.06 & 0.32 $\pm$ 0.05 & 0.29 $\pm$ 0.04 & 0.27 $\pm$ 0.04 & 0.25 $\pm$ 0.03 \\
& PSNR $\uparrow$ & 26.39 $\pm$ 3.01 & 20.23 $\pm$ 1.39 & 20.17 $\pm$ 1.22 & 20.12 $\pm$ 1.09 & 19.97 $\pm$ 0.92 & 19.68 $\pm$ 0.70 \\
\midrule\hline
\multirow{3}{*}{\textbf{sIQT}} 
& SSIM $\uparrow$ & 0.88 $\pm$ 0.02 & 0.88 $\pm$ 0.04 & 0.87 $\pm$ 0.03 &  0.87 $\pm$ 0.03 & 0.87 $\pm$ 0.03 & 0.86 $\pm$ 0.03 \\
& PSNR $\uparrow$ & 30.21 $\pm$ 3.50 & 26.22 $\pm$ 2.04 & 25.18 $\pm$ 2.18 & 24.62 $\pm$ 2.06 &  24.14 $\pm$ 2.19 & 23.96 $\pm$ 2.04 \\
\midrule\hline
\multirow{3}{*}{\textbf{kSURF}} 
& SSIM $\uparrow$ & \textbf{0.89 $\pm$ 0.03} & \textbf{0.88 $\pm$ 0.04} & \textbf{0.88 $\pm$ 0.03} & \textbf{0.88 $\pm$ 0.03} & \textbf{0.88 $\pm$ 0.03} & \textbf{0.88 $\pm$ 0.03} \\
& PSNR $\uparrow$ & \textbf{30.30 $\pm$ 3.48} & \textbf{26.45 $\pm$ 1.87} & \textbf{25.40 $\pm$ 2.12} & \textbf{24.80 $\pm$ 2.12} & \textbf{24.23 $\pm$ 2.08} & \textbf{24.01 $\pm$ 1.78} \\
\bottomrule
\end{tabular}
}
\end{subtable}
\end{table*}

\subsection{Qualitative Results using Synthetic LF-MRI Datasets}
\begin{figure*}
\centering
\includegraphics[width=0.8\textwidth]{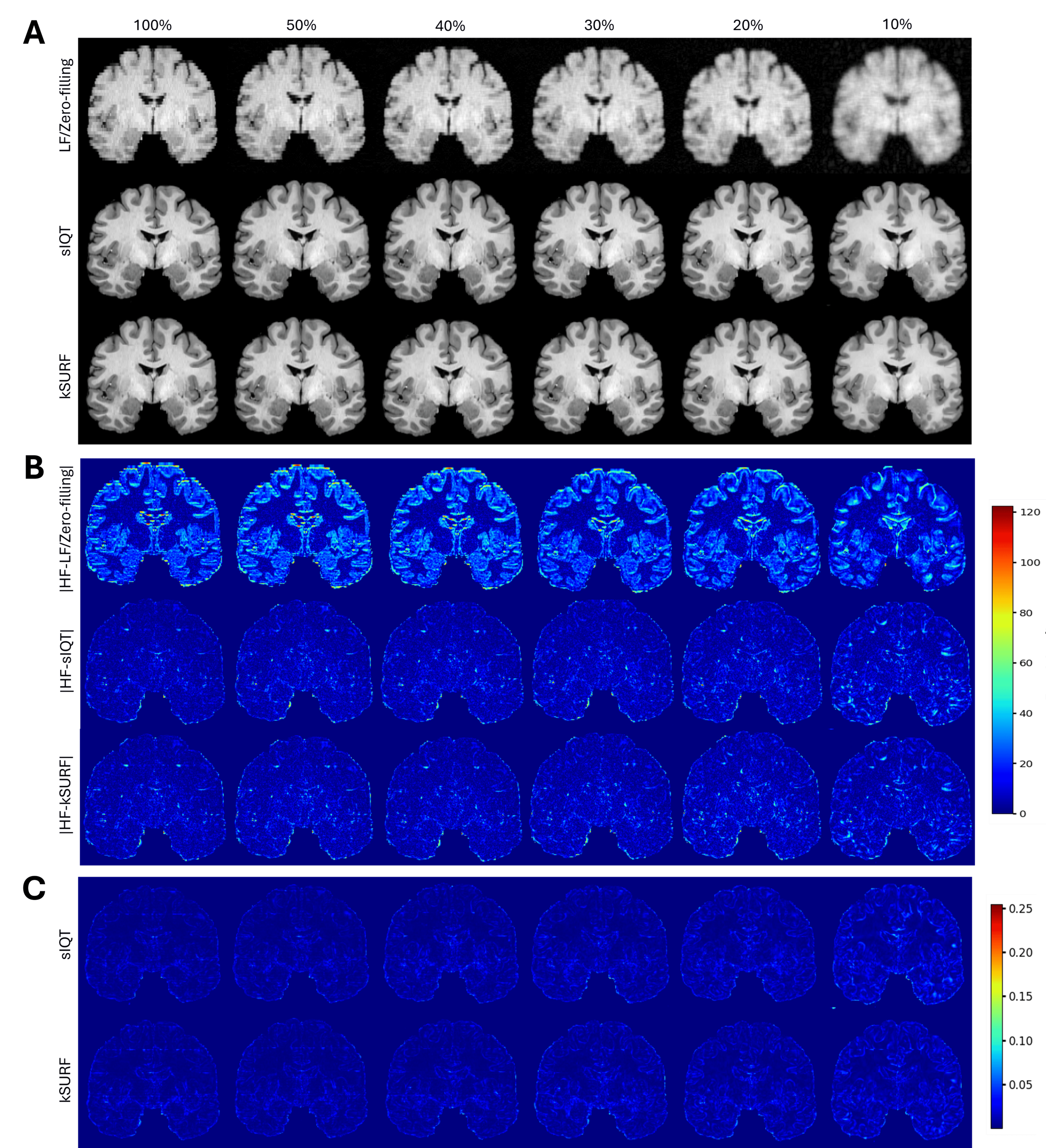}
\caption{Results using InD data. (A) reconstructions using LF/zero-filling, sIQT and kSURF across increasing pseudo-radial undersampling rates. (B) Corresponding error maps. (C) Corresponding uncertainty maps.} 
\label{fig:PRS_InD1(2)}
\end{figure*}

\begin{figure*}
\centering
\includegraphics[width=0.8\textwidth]{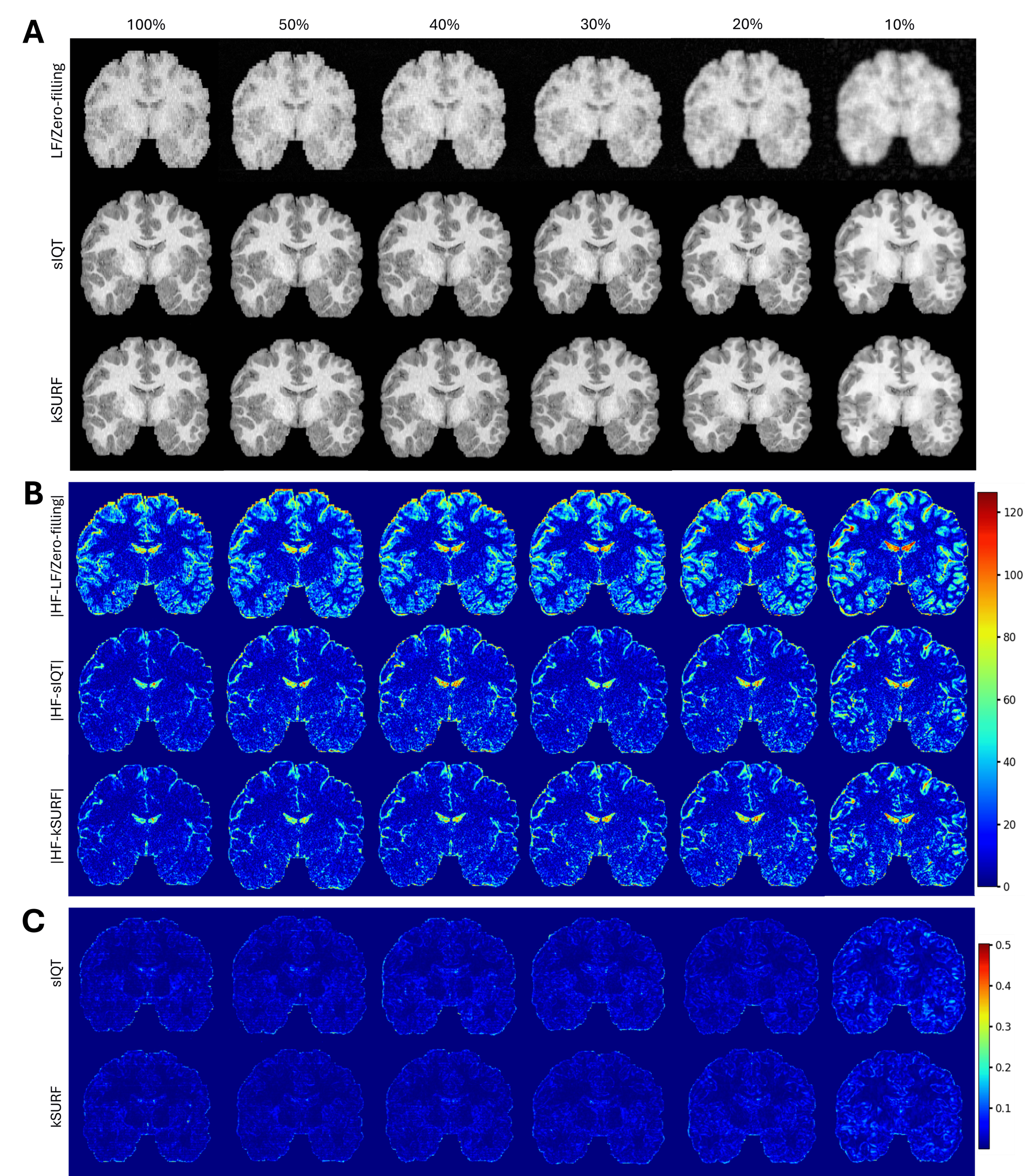}
\caption{Results using OOD data. (A) reconstructions using LF/zero-filling, sIQT and kSURF across increasing pseudo-radial undersampling rates. (B) Corresponding error maps. (C) Corresponding uncertainty maps.} 
\label{fig:PRS_OOD(2)}
\end{figure*}

To further evaluate the performance of the proposed approach, Figures \ref{fig:PRS_InD1(2)} and \ref{fig:PRS_OOD(2)} display example reconstructed brain test images from the HCP dataset, showing LF/zero-filling, and both spatial and k-space super-resolved images using the pseudo-radial sampling pattern at various undersampling ratios. The figures also show the error maps, highlighting the absolute differences between the original HF-MR images and their corresponding LF/zero-filling and reconstructed counterparts, as well as the corresponding uncertainty quantification maps, which provide valuable insights into the model's confidence across various undersampling ratios. The behavior with Cartesian and random sampling is similar, and therefore are not shown here. 

The qualitative results are consistent with the quantitative findings reported in Tables \ref{tab:InD1pcr} and \ref{tab:OODpcr}. As expected, LF/zero-filling reconstructions exhibit substantial visual degradation at lower sampling rates, characterised by pronounced blurring and loss of fine anatomical details. This degradation is particularly evident at the 10\% sampling level, where structural distortions and intensity inconsistencies become more pronounced. In contrast, both sIQT and kSURF consistently recover finer details, producing images with improved structural fidelity and reduced visual artifacts. Even under aggressive undersampling, both methods maintain a clear depiction of anatomical structures with minimal blurring. On the other hand, the residual error maps further corroborate these observations. LF/zero-filling reconstructions show large residual errors that increase as the undersampling rate becomes more severe. In comparison, both sIQT and kSURF yield significantly lower residual errors across all sampling rates. A direct comparison between sIQT and kSURF reveals that kSURF reconstructions exhibit consistently lower errors. On the other hand, kSURF demonstrates lower uncertainty than their spatial-domain super-resolved counterparts. This finding is consistent with the reduced residual errors observed in the corresponding error maps, further supporting the robustness of kSURF.

\begin{figure}
\centering
\includegraphics[width=0.38\textwidth]{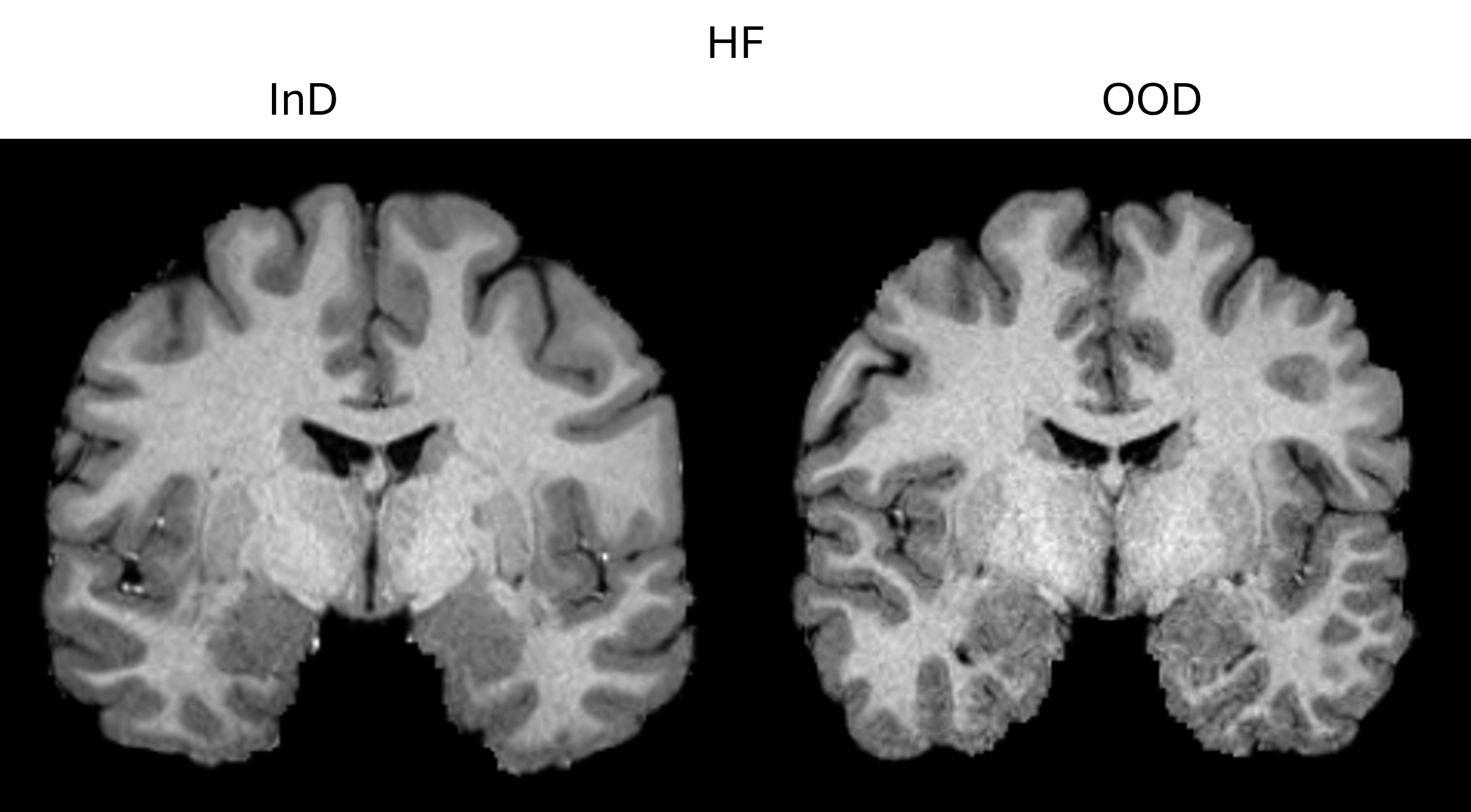}
\caption{Ground-truth high-field MR images used as representative examples for the InD and OOD test datasets.}
\label{fig:HFimg}
\end{figure}

\subsection{Comparison with existing work}
While the primary objective of this work is to reconstruct high-field–like MR images from under-sampled LF-MRI k-space and to compare this with spatial-domain super-resolution, we further assess the effectiveness of our reconstruction approach by benchmarking it against several existing methods. Specifically, we compare our method with low-rank total variation (LRTV) \cite{shi2015lrtv}, efficient sub-pixel convolutional neural network (ESPCN) \cite{shi2016real}, isotropic (ISO) U-Net \cite{cciccek20163d, lin2021generalised}, and 3D super-resolution U-Net (3D SR U-Net) \cite{heinrich2017deep, lin2019deep}. All comparison methods originally operate in the spatial domain and were therefore evaluated using the same input data and assessed relative to super-resolution performed on fully sampled (100\%) reconstructions. Table \ref{tab:EXT_METiqt} reports  PSNR and SSIM values for all methods using both InD and OOD datasets. The proposed kSURF approach consistently achieves the highest PSNR and SSIM scores, outperforming all baseline techniques. These results demonstrate that kSURF provides more robust and higher-quality reconstructions compared to existing spatial-domain methods.

\setlength{\arrayrulewidth}{0.5mm}
\setlength{\tabcolsep}{8pt}  
\begin{table}
\caption{Comparison with other existing methods (in the spatial domain) using 100\% sampling.}
\centering
\resizebox{\columnwidth}{!}{%
\begin{tabular}{ |c|c|c|c| }
\hline
\textbf{Methods} & \textbf{Metric} & \textbf{InD} & \textbf{OOD} \\
\hline

\multirow{2}{*}{\textbf{LF/Zero-filling}}
& SSIM  & 0.90 $\pm$ 0.01 & 0.84 $\pm$ 0.02 \\
& PSNR  & 29.95 $\pm$ 2.87 & 26.39 $\pm$ 3.01 \\
\hline

\multirow{2}{*}{\textbf{LRTV}}
& SSIM  & 0.85 $\pm$ 0.01 & 0.79 $\pm$ 0.02 \\
& PSNR  & 28.17 $\pm$ 2.87 & 24.60 $\pm$ 3.10 \\
\hline

\multirow{2}{*}{\textbf{ESPCN}}
& SSIM  & 0.92 $\pm$ 0.09 & 0.84 $\pm$ 0.03 \\
& PSNR  & 32.21 $\pm$ 4.33 & 27.10 $\pm$ 3.90 \\
\hline

\multirow{2}{*}{\textbf{ISO U-Net}}
& SSIM  & 0.94 $\pm$ 0.01 & 0.86$ \pm$ 0.02 \\
& PSNR  & 33.92 $\pm$ 2.96 & 28.90 $\pm$ 3.20 \\
\hline

\multirow{2}{*}{\textbf{3D SR U-Net}}
& SSIM  & 0.94 $\pm$ 0.01 & 0.87 $\pm$ 0.02 \\
& PSNR  & 33.88 $\pm$ 3.01 & 29.00 $\pm$ 3.20 \\
\hline

\multirow{2}{*}{\textbf{sIQT}}
& SSIM  & 0.96 $\pm$ 0.01 & 0.88 $\pm$ 0.02 \\
& PSNR  & 35.55 $\pm$ 2.54 & 30.21 $\pm$ 3.50 \\
\hline

\multirow{2}{*}{\textbf{kSURF}}
& SSIM  & \textbf{0.97 $\pm$ 0.01} & \textbf{0.89 $\pm$ 0.03} \\
& PSNR  & \textbf{35.60 $\pm$ 2.45} & \textbf{30.30 $\pm$ 3.48} \\
\hline
\end{tabular}}
\label{tab:EXT_METiqt}
\end{table}

\subsection{Results using real LF-MRI datasets and and clinical relevance}
\begin{figure*}
    \centering
\includegraphics[width=0.8\textwidth]{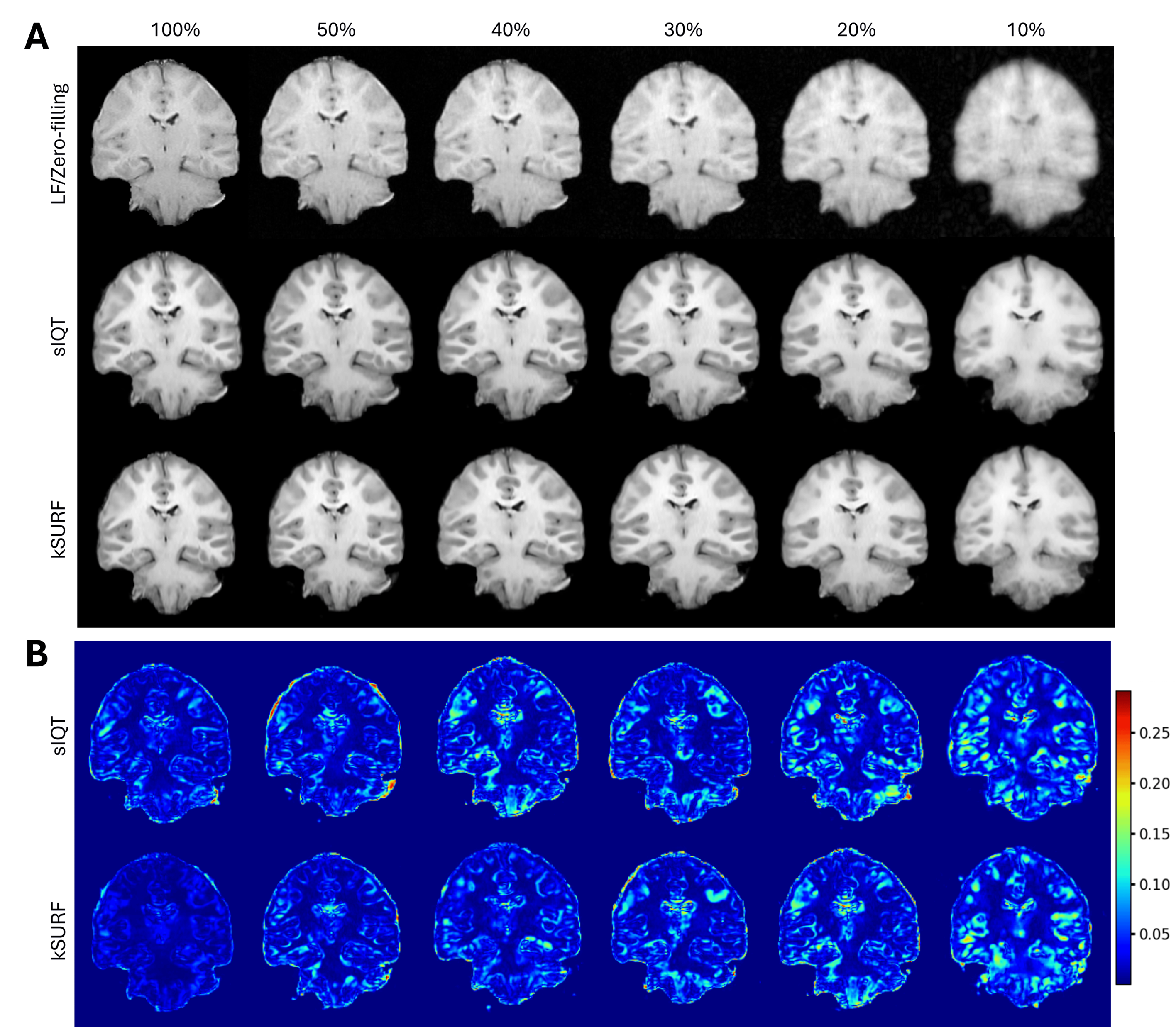}
    \caption{Results using real low-field data from healthy subjects. (A) reconstructions using LF/zero-filling, sIQT and kSURF across increasing pseudo-radial undersampling rates. (B) Corresponding uncertainty maps.}
    \label{fig:PRS_hT1w}
\end{figure*}

\begin{figure*}
    \centering
    \includegraphics[width=0.81\textwidth]{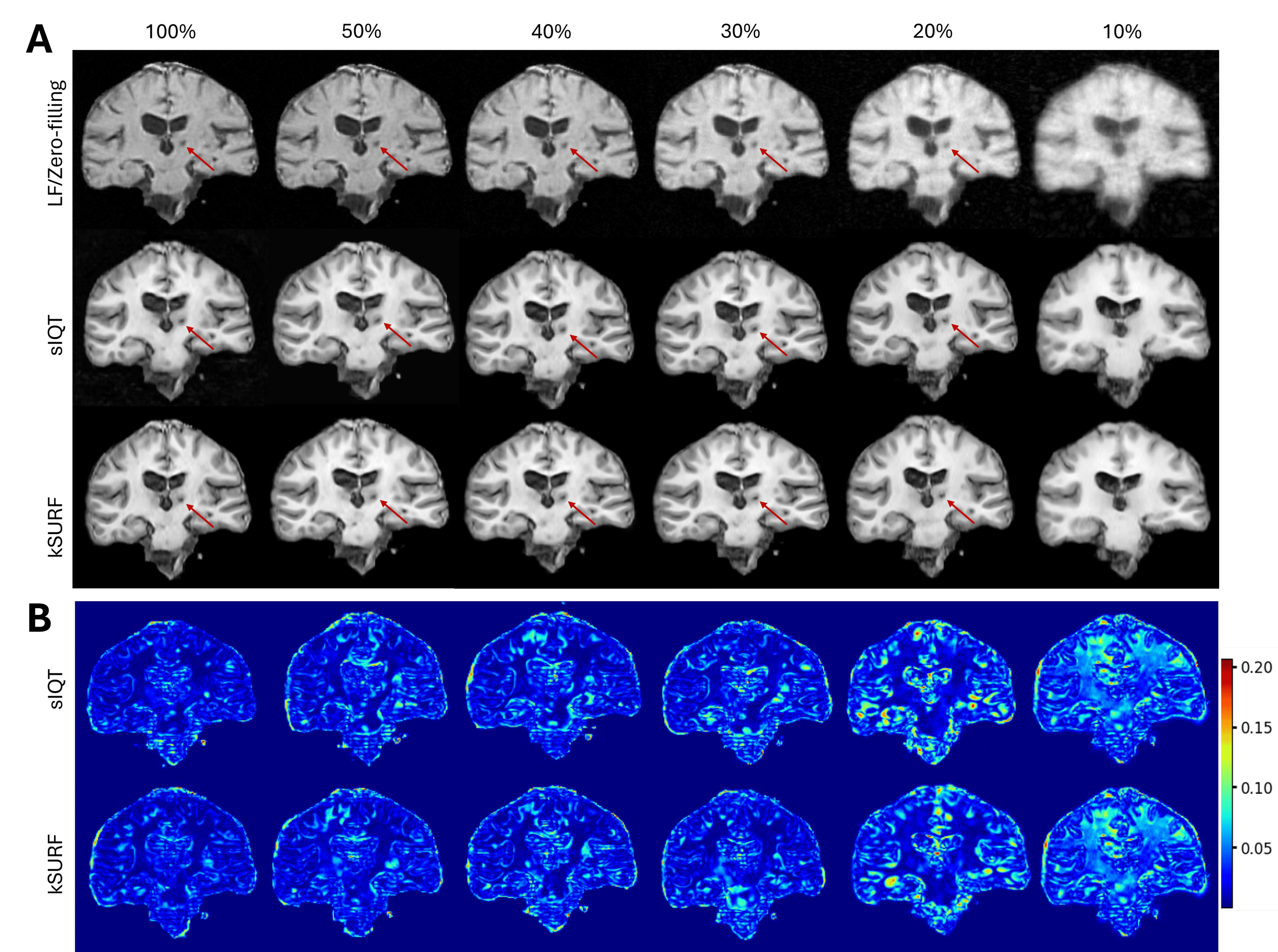}
    \caption{Results using real low-field data from unhealthy subjects. (A) reconstructions using LF/zero-filling, sIQT and kSURF across increasing pseudo-radial undersampling rates, with stroke pointed by a red arrow. (B) Corresponding uncertainty maps.}
    \label{fig:PRS_uT1w}
\end{figure*}

To further validate the proposed approach under realistic acquisition conditions, we evaluated our method on real LF-MRI data acquired using a 0.3T scanner \cite{wogu2025labeled}. The dataset included both healthy subjects and patients diagnosed with Parkinson’s disease, for whom clinically relevant stroke-related abnormalities were present. Figures \ref{fig:PRS_hT1w} and \ref{fig:PRS_uT1w} show reconstructed T1-weighted brain images obtained using LF/zero-filling, sIQT, and kSURF across increasing undersampling rates with a pseudo-radial sampling pattern. Corresponding uncertainty quantification maps are also provided. Reconstructions using Cartesian and random undersampling exhibited comparable qualitative trends and are therefore not shown.

For both healthy and pathological cases, LF/zero-filling reconstructions exhibit progressive degradation as the sampling rate decreases, manifested by increased blurring and loss of fine anatomical detail. In contrast, both sIQT and kSURF consistently recover finer structures with substantially fewer artefacts across all undersampling levels. This improvement is evident in the preservation of key anatomical contrasts, including grey matter, white matter, and cerebrospinal fluid, even under aggressive undersampling. The uncertainty quantification maps further corroborate these observations, revealing increased uncertainty as undersampling becomes more severe. Notably, kSURF consistently exhibits lower uncertainty compared to sIQT, indicating more stable and reliable reconstructions. These findings are consistent with the trends observed in the synthetic InD and OOD experiments reported earlier.

From a clinical perspective, the advantages of kSURF are most pronounced in the pathological case, as illustrated in Figure \ref{fig:PRS_uT1w} (the figure is enlarged for better visualisation of the clinical features). Stroke-related features that are poorly defined or obscured in the low-field and zero-filled reconstructions become clearly discernible in the super-resolved images produced by both sIQT and kSURF. These pathological signs remain visible across most undersampling rates. At the most aggressive undersampling level (10\%), the visibility of these features is reduced for both LF/zero-filling and super-resolved reconstructions, due to the excessively severe sampling.

Overall, these results demonstrate that LF-MRI acquisitions can be substantially accelerated through k-space undersampling while preserving clinically relevant image information. By operating directly in k-space, the proposed framework achieves improved structural fidelity and lower reconstruction error and uncertainty compared with spatial-domain approaches, particularly under aggressive undersampling. The consistent preservation of anatomical detail and pathological features across a wide range of sampling rates highlights the robustness of the method and supports k-space–based super-resolution as an effective strategy for accelerating low-field MRI without severe degradation of image quality.

To evaluate its effectiveness, we test kSURF across a range of undersampling scenarios, where it consistently outperforms existing methods, achieving high-field MR-like image quality even under aggressive k-space undersampling. Quantitative analysis demonstrates that kSURF more effectively restores valuable frequency-domain information, preserving fine anatomical details and voxel-level fidelity across multiple sampling patterns. These findings are further supported by qualitative evaluations, which reveal enhanced tissue contrast, reduced blurring, and improved structural consistency, in alignment with the quantitative metrics. 

Error and uncertainty quantification maps further corroborate these observations, showing lower reconstruction errors and more stable confidence estimates. Importantly, kSURF maintains robust performance under domain shift, generalising well to out-of-distribution data, emphasising its reliability and adaptability. This robustness extends to real-world low-field MRI acquisitions from both healthy and pathological subjects, where kSURF successfully recovers clinically relevant anatomical and pathological features, such as stroke-related abnormalities, highlighting its potential to substantially accelerate low-field MRI acquisitions without compromising diagnostic quality.

Beyond reconstruction accuracy, kSURF exhibits characteristics that are important for practical use in low-field MRI settings. By enabling reliable frequency-domain information recovery from substantially undersampled k-space data, the framework supports meaningful reductions in acquisition time, an important consideration in time-constrained and resource-limited clinical environments. In addition, the incorporation of uncertainty quantification enhances interpretability by providing confidence estimates that allow reconstructed images to be assessed in context and support downstream clinical analysis. While the present results demonstrate the feasibility and clinical promise of this approach, future work will focus on further integration of uncertainty information into acquisition strategies. In particular, adaptive k-space sampling strategies informed by uncertainty estimates may enable more efficient use of acquired data, improving reconstruction reliability while minimising acquisition demands, thereby further extending the clinical relevance of the proposed framework.

\section{Conclusion}
\label{sec:Conclusion}
In this work, we proposed kSURF, a novel k-space–based image super-resolution/quality transfer framework that jointly reconstructs super-resolved MRI images from undersampled low-field k-space data, in contrast to conventional approaches that treat reconstruction and enhancement as separate tasks. By leveraging k-space information rather than relying on spatial-domain post-processing, the proposed method enables k-space undersampling, thereby accelerating scan times while maintaining high diagnostic image quality. Experiments on low-field brain MRI datasets demonstrate that k-space–driven image super-resolution/quality transfer outperforms spatial-domain counterparts while enabling significant scan-time acceleration, as undersampled k-space reconstructions achieve quality comparable to fully sampled acquisitions. In addition, uncertainty quantification is incorporated to assess the effects of k-space undersampling and reconstruction reliability, providing confidence maps that enhance interpretability and clinical trust. Future work will explore adaptive k-space sampling strategies guided by uncertainty maps to further optimise acquisition efficiency while maintaining high-quality reconstructions with minimal data.




\bibliographystyle{ieeetr}
\bibliography{reference}

\begin{IEEEbiography}[{\includegraphics[width=1.11in,height=1.25in]{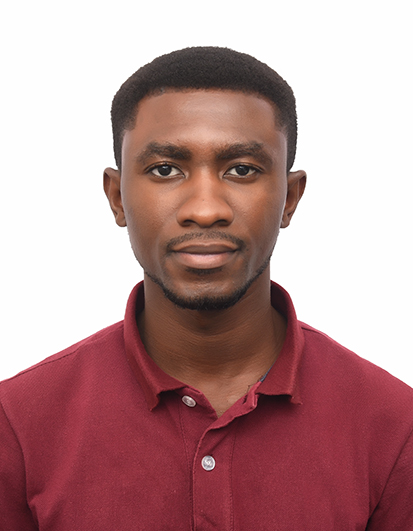}}]{Daniel Tweneboah Anyimadu} is a PhD candidate in Computer Science at the University of Exeter (UoE), specialising in AI-driven diagnostic solutions with a particular focus on low-field Magnetic Resonance Image (MRI) reconstruction from undersampled k-space data to enhance medical imaging. Before embarking on his doctoral journey at UoE, he earned his MSc in Medical Imaging and Applications (MAIA) through the esteemed Erasmus+ programme. \end{IEEEbiography}

\begin{IEEEbiography}[{\includegraphics[width=1.11in,height=1.25in]{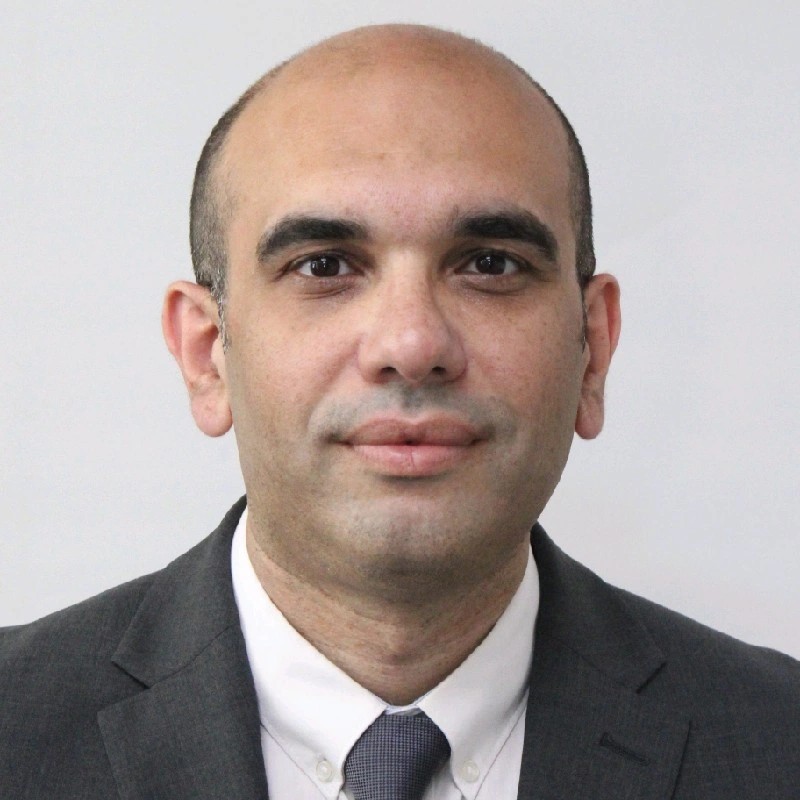}}]{Mohamed Abdalla} received the M.B.B.Ch. degree from Menoufia University Faculty of Medicine, Egypt, graduating with Excellence and Honours and ranking second in his class. He obtained the MSc degree in Surgery from Menoufia University in 2009, followed by the MRCS qualification in England in 2016 and the Fellowship of the Royal College of Surgeons of England in Neurosurgery (FRCS-SN) in 2019. He has held academic and clinical neurosurgical appointments in Egypt, Saudi Arabia, and the United Kingdom, including roles at Imperial College Healthcare NHS Trust, and St George’s University Hospitals NHS Foundation Trust. He is currently a Consultant in neurosurgery at the Royal Devon and Exeter Hospital, United Kingdom.
\end{IEEEbiography}

\begin{IEEEbiography}[{\includegraphics[width=1.11in,height=1.25in]{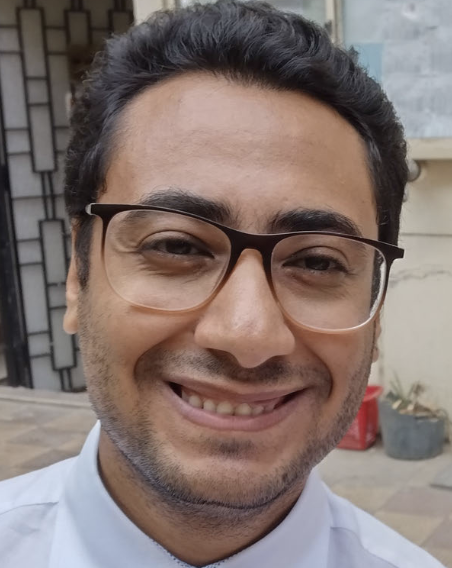}}]{Mohamed M. Abdelsamea} received the Ph.D. (with Doctor Europaeus) degree in computer science and engineering from the IMT-Institute for Advanced Studies, Lucca, Italy, in 2015. He is currently a Senior Lecturer in computer science (machine learning and computer vision) with the Computer Science Department, University of Exeter; and a fellow of the British Higher Education Academy (HEA). Before joining Exeter University, he was a Senior Lecturer in data and information science with the School of Computing and Digital Technology, Birmingham City University, where he was also a Leading Member of the Computer Vision Theme within the DAAI Research Group. He has been working on developing artificial intelligence tools for healthcare and life science applications, since 2005. He has been with several groups of biologists, pathologists, and computer scientists in different research projects in Egypt, Singapore, Italy, and the U.K. He was with the School of Computer Science, Nottingham University; Mechanochemical Cell Biology, Warwick University; Nottingham Molecular Pathology Node (NMPN) and the Division of Cancer and Stem Cells, Nottingham Medical School, as a Research Fellow. In 2016, he received the prestigious Marie-Cuire Research Fellowship from the School of Computer Science, Nottingham University. His current research interests include the development of statistical/classical machine learning and deep learning approaches in the areas of medical image segmentation, medical image classification, semantic/instance segmentation, feature extraction, and data mining and machine learning, with the overall ambition to assist human investigation.
\end{IEEEbiography}

\begin{IEEEbiography}[{\includegraphics[width=1.11in,height=1.25in]{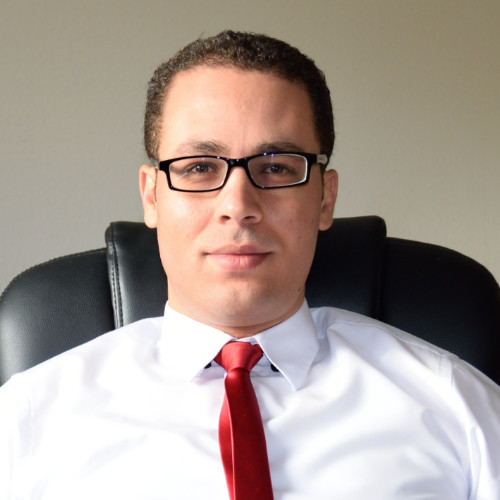}}]{Ahmed Karam Eldaly} achieved an Erasmus Mundus joint Masters degree in Computer Vision and Robotics (VIBOT) from Universite de Bourgogne in France, Universitat de Girona in Spain, and Heriot-Watt University (HWU) in the UK. This programme was funded by the European Commission, and he completed it with a Distinction in 2015. In October 2018, he was awarded the PhD in Electrical, Electronic and Computer Engineering jointly with School of Engineering, HWU, and The Queens Medical Research Institute, The University of Edinburgh (UoE). He worked as a Research Fellow in computational imaging at HWU, UoE, and University College London (UCL). He is currently a Lecturer (Assistant Professor) in Computer Vision and AI for Health at the University of Exeter, and an Honorary Lecturer at the Department of Computer Science, UCL. His primary research interests encompass computational imaging, machine learning, statistical signal, and image processing. \end{IEEEbiography}

\EOD

\end{document}